\documentclass[11pt]{article}
\usepackage{coling2018}
\usepackage{times}
\usepackage{url}
\usepackage{latexsym}
\usepackage{graphicx}
\usepackage{subcaption}
\graphicspath{ {images/} }
\usepackage[ruled,vlined,linesnumbered]{algorithm2e}
\usepackage{amsmath}
\usepackage{multirow}
\usepackage{booktabs}
\usepackage[title]{appendix}
\usepackage{array, makecell}%

\DeclareMathOperator*{\argmax}{argmax}

\title{Character-Level Feature Extraction\\
	with Densely Connected Networks
}

\author{Chanhee Lee\textsuperscript{1}, Young-Bum Kim\textsuperscript{2}, Dongyub Lee\textsuperscript{1}, HeuiSeok Lim\textsuperscript{1}\thanks{\hspace{0.1cm} corresponding author} \\
  \textsuperscript{1}Korea University, Republic of Korea \\
  {\tt \{chanhee0222, judelee93, limhseok\}@korea.ac.kr} \\
  \textsuperscript{2}Amazon Alexa \\
  {\tt youngbum@amazon.com} \\}

\date{}

\begin{document}
\maketitle
\begin{abstract}
	
  Generating character-level features is an important step for achieving good results in various natural language processing tasks. To alleviate the need for human labor in generating hand-crafted features, methods that utilize neural architectures such as Convolutional Neural Network (CNN) or Recurrent Neural Network (RNN) to automatically extract such features have been proposed and have shown great results. However, CNN generates position-independent features, and RNN is slow since it needs to process the characters sequentially. In this paper, we propose a novel method of using a densely connected network to automatically extract character-level features. The proposed method does not require any language or task specific assumptions, and shows robustness and effectiveness while being faster than CNN- or RNN-based methods. Evaluating this method on three sequence labeling tasks - slot tagging, Part-of-Speech (POS) tagging, and Named-Entity Recognition (NER) - we obtain state-of-the-art performance with a 96.62 F1-score and 97.73\% accuracy on slot tagging and POS tagging, respectively, and comparable performance to the state-of-the-art 91.13 F1-score on NER.

\end{abstract}

\section{Introduction}

%
%
\blfootnote{
	%
	%
	%
	%
	%
	%
	\hspace{-0.65cm}  
	This work is licensed under a Creative Commons 
	Attribution 4.0 International License.
	License details:
	\url{http://creativecommons.org/licenses/by/4.0/}
}

Effectively extracting character-level features from words is crucial in many Natural Language Processing (NLP) tasks, such as Named Entity Recognition (NER), Part-of-Speech (POS) tagging, and Slot tagging. Thus, most state-of-the-art methods for these tasks exploit some kind of character-level features \cite{huang2015bidirectional,sarikaya2007joint,kim2011universal,kim2012universal,dos2014learning}. Recently, generating character-level features with neural architectures such as Convolutional Neural Network (CNN) or Recurrent Neural Network (RNN) has drawn much attention, mainly because it doesn't require human labor and shows superior performance \cite{ma2016end,dos2014learning}. However, CNN struggles at distinguishing anagrams, and RNN is inherently slow due to its sequential nature. \\
In this paper, we propose an effective and efficient way of extracting character-level features using a densely connected network. The key benefits of the proposed method can be summarized as follows. First, it does not require any hand-crafted features or data preprocessing. Each word is processed based on n-gram statistics of the training data, and vectorized using bag-of-characters. Additional features are based on hexadecimal values of the character-set (e.g. UTF-16) and number of characters in the word. Second, it extracts effective character-level features while being efficient. State-of-the-art performance can be achieved using this method, and the feature extraction is done with a simple densely connected network with a single hidden layer. Third, it doesn't depend on features that are language or task specific, such as character type features or gazetteer (i.e. lists of known named entities such as cities or organization names). The only requirement for adopting this method is that the language should be processable as a sequence of words, which is made of sequence of characters. These benefits, combined with minimum requirements for application, make the proposed method an easy replacement for conventional methods such as CNN or RNN.\\
Our contributions are three-fold: 1) We propose an effective yet efficient method for character-level feature extraction; 2) We quantitatively show that the proposed method is superior to CNN and RNN via extensive evaluation; 3) We achieve state-of-the-art or comparable to state-of-the-art performance on three of the most popular and well-studied sequence tagging tasks - Slot tagging, Part-of-Speech (POS) tagging, and Named Entity Recognition (NER).

\section{Related Work}
Prior to the introduction of neural architectures for character-level feature generation, manually engineered features were designed by experts based on language and/or domain knowledge. One example is word shape, in which each word is mapped to a simplified representation that encodes information such as capitalization, numerals, and length (e.g. CoNLL-2003 to AaAAA-0000). \newcite{finkel2005exploring} combined this feature with other information such as n-grams and gazetteers to train a conditional Markov model for identification of gene and protein names in biomedical documents. \newcite{huang2015bidirectional} introduced more hand-crafted features utilizing punctuation or non-letters and used these as an input to a Bi-LSTM-CRF tagger for POS tagging, CoNLL-2000 chunking, and CoNLL-2003 NER. Even though these kinds of hand-crafted features showed strong empirical results, they are more expensive than our approach in that they require expert knowledge of the target domain and language.

In recent years, methods that utilize neural networks to automatically extract character-level features have been proposed. The most widely adopted and successful method for this is CNN. \newcite{dos2014learning} combined this approach with a window-based fully-connected neural network tagger to perform English and Portuguese POS tagging. This work achieved state-of-the-art results in Portuguese and near state-of-the-art results in English. In \newcite{ma2016end}, a Bi-LSTM-CRF model incorporated with a character-level CNN is trained in an end-to-end fashion. They evaluated this approach on English POS tagging and NER, achieving state-of-the-art performance on both tasks. However, feature vectors generated by CNN are position-independent due to the max-over-time pooling layer, and are more sensitive to model weight initialization compared to the method proposed in this paper.

Another effective way of generating feature vectors from a variable length sequence of characters is to use RNN. For instance, \newcite{lample2016neural} extracted character-level features using a bi-directional LSTM and used them with pre-trained word embeddings as word representations for another Bi-LSTM-CRF model. Evaluating this model for NER, they obtained state-of-the-art results for Dutch, German, and Spanish, and close to state-of-the-art results for English. Intuitively, character-level feature generation via RNN should be more effective than CNN, since RNN processes each character sequentially and thus should form a better model of character ordering. However, \newcite{reimers2017reporting} empirically showed that these two methods have no statistically significant difference in terms of performance. Furthermore, RNN has a higher time-complexity caused by its sequential nature, which makes it less favorable.

\section{Proposed Method}
The proposed method is built on bag-of-characters (BOC) representation. However, BOC is prone to anagrams and thus is susceptible to word collisions, i.e. different words having the same vector representation. The main focus of the proposed method is to minimize word collision while maintaining the key benefits described above. To achieve this goal, we split the word into $k$ pieces, and each piece is vectorized using BOC. Then, two non hand-crafted-features are extracted from the word - character order and word length. These sparse vectors are concatenated and normalized to form the sparse character-level feature vector. For a n-dimensional vector $\mathbf{x}=\left< x_1, x_2, \cdots, x_n \right>$, normalizing is done as follows:

\begin{equation}
x_i'=\frac{x_i}{\sum_{j=1}^n x_j}
\end{equation}

\begin{figure}
	\includegraphics[width=\columnwidth]{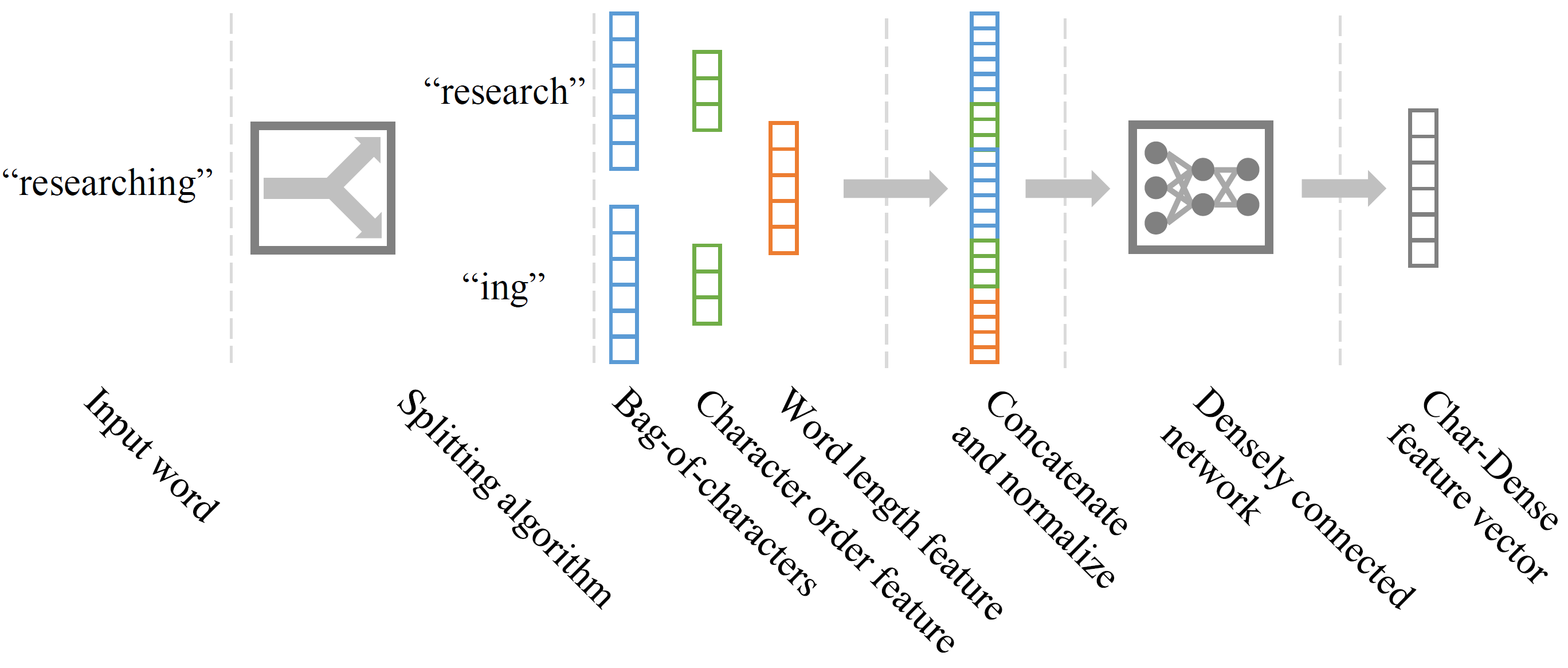}
	\centering
	\caption{Process of generating the character-level feature vector of a word using the proposed method.}
	\label{fig:char_dense_process}
\end{figure}

This sparse vector is then fed into a densely connected network with a single hidden layer to obtain the final dense character feature vector. Note that the sparse vector representation of each word is fixed, so it can be cached for efficiency. Figure~\ref{fig:char_dense_process} illustrates the overall process.

\subsection{Splitting Words}

Each word is split into $k$ pieces to reduce the number of word collisions. To maintain the ordering of pieces, concatenation is used instead of summation or averaging to merge the vectors. Word splitting is done based on n-gram frequency. First, n-gram statistics $ C_{ng} $ is collected from the training corpus where $ C_{ng}(s) $ is the number of times the n-gram $s$ appears in the corpus. Then, the n-gram with the highest frequency gets merged into a single piece, and this merging is repeated until only $k$ n-grams are left. The number of pieces $k$ per word is a configurable hyperparameter. Finally, each piece is converted into a fixed length vector using BOC. The detailed algorithm is presented in Algorithm~\ref{splitting_algorithm}. This process is similar to the byte-pair encoding method in \newcite{sennrich2015neural}, except that in the proposed method each word can only be split into $k$ pieces whereas byte-pair encoding produces an arbitrary number of pieces. Producing a fixed number of pieces is important, since concatenation is used to merge the vectors.

\begin{algorithm}
	\caption{Splitting word into $ k $ pieces}
	\label{splitting_algorithm}
	\DontPrintSemicolon
	\SetAlgoLined
	\SetKwInOut{Input}{Input}\SetKwInOut{Output}{Output}
	\Input{word $ w=(c_1, c_2, \cdots, c_n) $, n-gram statistics $ C_{ng} $, number of pieces $ k $}
	\Output{ $S = (s_1, \cdots, s_k) $ where $ s_1+s_2+\cdots+s_k = w $}
	\BlankLine
	$ S \leftarrow w $ \\
	\While{$ |S| > k $} {
		$ \displaystyle m = \argmax_i C_{ng}(c_i + c_{i+1}) $ \\
		$ S \leftarrow (\cdots, s_{m-1}, s_m + s_{m+1}, s_{m+2}, \cdots) $ \\
	}
	\While{$ |S| < k $} {
		Append empty string to $ S $
	}
	\Return $ S $
\end{algorithm}

\subsection{Character Order Feature}
Every character that has a digital representation can be converted into a numerical value via some character-set (e.g. UTF-16). Then, it is possible to numerically compare two characters. Let $ T=\{c_1, c_2, \cdots, c_n\} $ be a character sequence of length $n$. Then $ F_{asc}(T, k) $, $ F_{des}(T, k) $, $ C_{asc}(T) $, and $ C_{des}(T) $ are defined as follows:\\

\begin{equation}
F_{asc}(T, k) =
\begin{cases}
	 1, \text{if } c_k < c_{k+1} \\
	 0, \text{otherwise}
\end{cases}
,\quad F_{des}(T, k) =
\begin{cases}
1, \text{if } c_k > c_{k+1} \\
0, \text{otherwise}
\end{cases}
\end{equation}

\begin{equation}
C_{asc}(T) = \sum_{k=1}^{n-1} F_{asc}(T, k)
\quad,\qquad C_{des}(T) = \sum_{k=1}^{n-1} F_{des}(T, k)
\end{equation}

Bi-grams with the same character repeating are ignored. A sequence of characters can then be categorized into one of three classes: $ C_{asc}(T) > C_{des}(T) $, $ C_{asc}(T) = C_{des}(T) $, $ C_{asc}(T) < C_{des}(T) $. This category info is calculated for each word piece, which is then converted into a 3-dimensional vector using one-hot encoding and concatenated to the sparse word piece vector.

\subsection{Word Length Feature}
To further reduce the number of word collisions, information about the word's length is added into the model. One-hot encoding is used to store an integer from 0 to 20, and any word exceeding 20 characters is treated as being 20 characters long.

\section{Model}
\begin{figure}
	\includegraphics[width=\columnwidth]{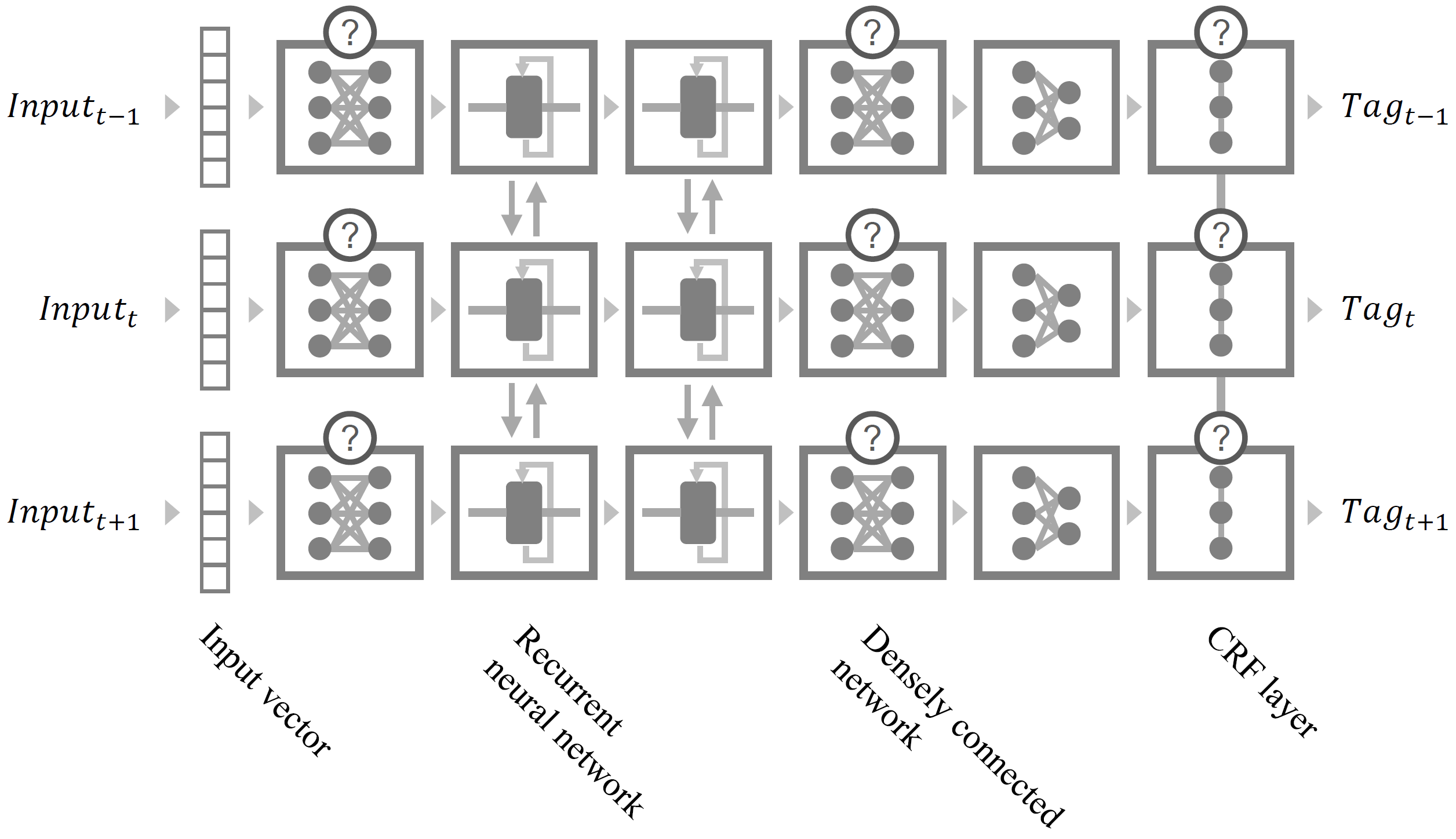}
	\centering
	\caption{Overview of model architecture for sequence tagging experiments. Question mark indicates that the component is optional.}
	\label{fig:model_overview}
\end{figure}

In this section, we describe the sequence tagging model's architecture in detail. Figure~\ref{fig:model_overview} illustrates the model architecture.

\subsection{Sequence Tagging with Bidirectional RNN}
In sequence tagging tasks, such as POS tagging or NER, both future and past input tokens are available to the model. Bidirectional RNNs~\cite{graves2005framewise} can efficiently make use of future and past features over a certain time frame. We use Long Short-Term Memory (LSTM) \cite{hochreiter1997long} for our RNN cell, which is better at capturing long-term dependencies than vanilla RNN \cite{kim2016frustratingly}. Output of the forward and backward RNN layers are summed to form the feature vector of each time-step. Each word is tagged based on this feature vector, using either a softmax layer or CRF layer.
To capture a more abstract and higher-level representation in different layers, a densely connected layer can be added before and after the Bi-LSTM layers. The input to this network at each time-step is the concatenation of the character-level feature vector and a pre-trained word vector (described in section \ref{section:training}).

\subsection{Conditional Random Field}
Even though a Bi-LSTM layer can efficiently extract features for each time-step utilizing past and future inputs, the prediction is made on each time-step, independent of past and future tag outputs. The Conditional Random Field (CRF) layer overcomes this limitation by considering state transition probability, thereby decoding the most probable output tag sequence \cite{kim2014training,kim2015new}. It has been shown that adding a CRF layer on top of a Bi-LSTM network can lead to statistically significant performance increases \cite{reimers2017reporting}. We also test a variant of our model using CRF as the final layer to perform tag sequence prediction.

\subsection{Stacking RNNs with Residual Connection}
Increasing the depth of the neural network architecture has proven to be an effective way of improving performance. However, naively stacking layers can lead to adversarial effects due to the degradation problem. Residual connection \cite{he2016deep} has shown to be an effective way to tackle this issue by creating a shortcut between layers. The same strategy is adopted to our model when there are more than one Bi-LSTM layers, in which case the input is added to the Bi-LSTM layer's output. 

\subsection{Dropout}
Dropout is a popular and effective way of regularizing neural network models, by randomly dropping nodes \cite{srivastava2014dropout}. In our model, Inverted dropout is applied to all densely connected layers for regularization. For the Bi-LSTM layers, variational recurrent dropout \cite{gal2016theoretically} is used, since naive dropout can deteriorate performance. The word embedding matrix is regularized using the method proposed in \newcite{gal2016theoretically}, i.e. dropping words at random.

\section{Training Details}
\label{section:training}
\textbf{Pre-trained Word Embeddings}
\space\space Utilizing word embeddings pre-trained on large unlabeled text has shown to be one of the most effective ways to increase performance on various NLP tasks. Our model uses the GloVe \cite{pennington2014glove} 300-dimensional vectors trained on the Common Crawl corpus with 42B tokens as word level features, as this resulted in the best performance in preliminary experiments. Words that do not appear in the training data are replaced with a special Out-of-Vocabulary (OOV) token. To train the vector of this token, we randomly swap words with OOV tokens while training with a 0.01 probability, as in \newcite{lample2016neural}. The word vector is then concatenated with the character-level feature vector and fed into the subsequent layer.

\noindent\textbf{Freezing Embeddings}
\space\space It is common practice to fine-tune the pre-trained word vectors through the training process. However, preliminary experiments have revealed that fine-tuning the word vectors results in lower performance than freezing the vectors, especially in the early stages of training. We hypothesize that randomly initialized weights in the model act as noise and degrade the pre-trained word vectors. To circumvent this issue, the embeddings are frozen for the first $ T_{freeze} $ phase of training so that they are not affected by untrained weights. We use $ T_{freeze} =$ 20\% for all experiments.

\noindent\textbf{Dynamic Batch Size}
\space\space \newcite{keskar2016large} showed that small batch sizes lead to more global and flat minimizers, while large batch sizes lead to more local and sharp minimizers. Therefore, starting from a small batch size and increasing it during training would result in a more global, but sharp minimizer. While having similar effect to learning rate decay, this strategy also has a benefit of accelerating the training as the batch size grows \cite{smith2017don}. Adopting this method, we start from a fixed initial batch size, and increase the batch size by a factor of two on each quarter of the course of training.

\noindent\textbf{Tagging Scheme}
\space\space It is reported that more complicated tagging schemes such as IOBES does not have statistically significant advantage over BIO scheme \cite{reimers2017reporting}, thus we adopt the BIO scheme for all experiments.

\noindent\textbf{Parameter Optimization}
\space\space Our network is trained by minimizing the cross entropy loss over the tags for the softmax model, or maximizing the log-likelihood of the tag sequence for the CRF model. The objective function is optimized using the gradient-based optimization algorithm Adam~\cite{kingma2014adam}. For all experiments, we implement the model using the TensorFlow~\cite{abadi2016tensorflow} library.

\noindent\textbf{Hyperparameter Tuning}
\space\space Most hyperparameters, with the following exceptions, are tuned on the development sets. Hyperparameters of the character-CNN and character-RNN models are adopted from \newcite{ma2016end} and \newcite{lample2016neural}, respectively. The chosen hyperparameters for all experiments are summarized in Appendix~\ref{appendix:hyperparameters}.

\section{Evaluation}
We evaluate the effectiveness of the proposed method using three of the most well-studied and common English sequence tagging tasks - Slot tagging, POS tagging, and NER. Note that to test the generalizability of the proposed method, we do not perform any preprocessing for all experiments. Details on each task and baseline models are described in this section. Table~\ref{tab:corpus_stats} summarizes the statistics of each task. 

\subsection{Slot Tagging}
For slot tagging, we use the Airline Travel Information System (ATIS) dataset. This dataset has 84 types of slot labels and 127 possible tags with BIO tagging scheme. Since this corpus lacks a development set, 20\% of the training data is randomly sampled and used as the development set for tuning the hyperparameters. This task's performance is measured in F1-score, which is calculated using the publicly available \textit{conlleval.pl} script.

\subsection{Part-of-Speech Tagging}
For POS tagging, we use the Wall Street Journal (WSJ) portion of the Penn TreeBank dataset \cite{marcus1993building} and adopt the standard split for part-of-speech tagging experiments - section 0-18 as training data, section 19-21 as development data, and section 22-24 as test data. This dataset contains 45 different POS tags. Model performance is measured by token-level accuracy.

\subsection{Named Entity Recognition}
For NER, the English portion of the CoNLL-2003 shared task \cite{tjong2003introduction} is used for evaluation. This dataset contains four different types of named entities, which results in nine possible tags with BIO tagging scheme and an 'O' tag. Like slot tagging, the final performance is measured in F1-score using the same \textit{conlleval.pl} script.

\begin{table}[t]
	\centering
	\begin{tabular}{@{}ccccccc@{}}
		\toprule
		\multirow{2}{*}{\textbf{Dataset}} & \multicolumn{2}{c}{\textbf{ATIS}} & \multicolumn{2}{c}{\textbf{PTB WSJ}} & \multicolumn{2}{c}{\textbf{CoNLL2003}} \\ \cmidrule(l){2-7} 
		& Sentences         & Tokens        & Sentences          & Tokens          & Sentences           & Tokens           \\ \midrule
		Training                          & 4978              & 56591         & 38219              & 912344          & 14987               & 204567           \\
		Develop.                          & -                 & -             & 5527               & 131768          & 3466                & 51578            \\
		Test                              & 893               & 9198          & 5462               & 129654          & 3684                & 46666            \\ \bottomrule
	\end{tabular}
	\caption{Corpus statistics of each task.}
	\label{tab:corpus_stats}
\end{table}

\subsection{Baseline Models}
Character-level CNN and character-level RNN are the most effective and widely adopted methods for character-level feature extraction, and thus are suitable as strong baseline methods. We implement these two methods to use them as baselines for comparison. The CRF layer has the effect of making the model robust to architectural differences \cite{reimers2017reporting}. Since the goal of baseline experiments is to evaluate the effect of difference in character-level feature generation methods, we use the softmax layer instead of the CRF layer for these experiments. Every aspect of the sequence tagging model except the character-level feature generation method is identical for all baseline experiments.

\section{Results and Discussion}
\subsection{Experimental Results}
\begin{figure}[t]
	\centering
	\begin{subfigure}[b]{0.3\textwidth}
		\includegraphics[width=\textwidth]{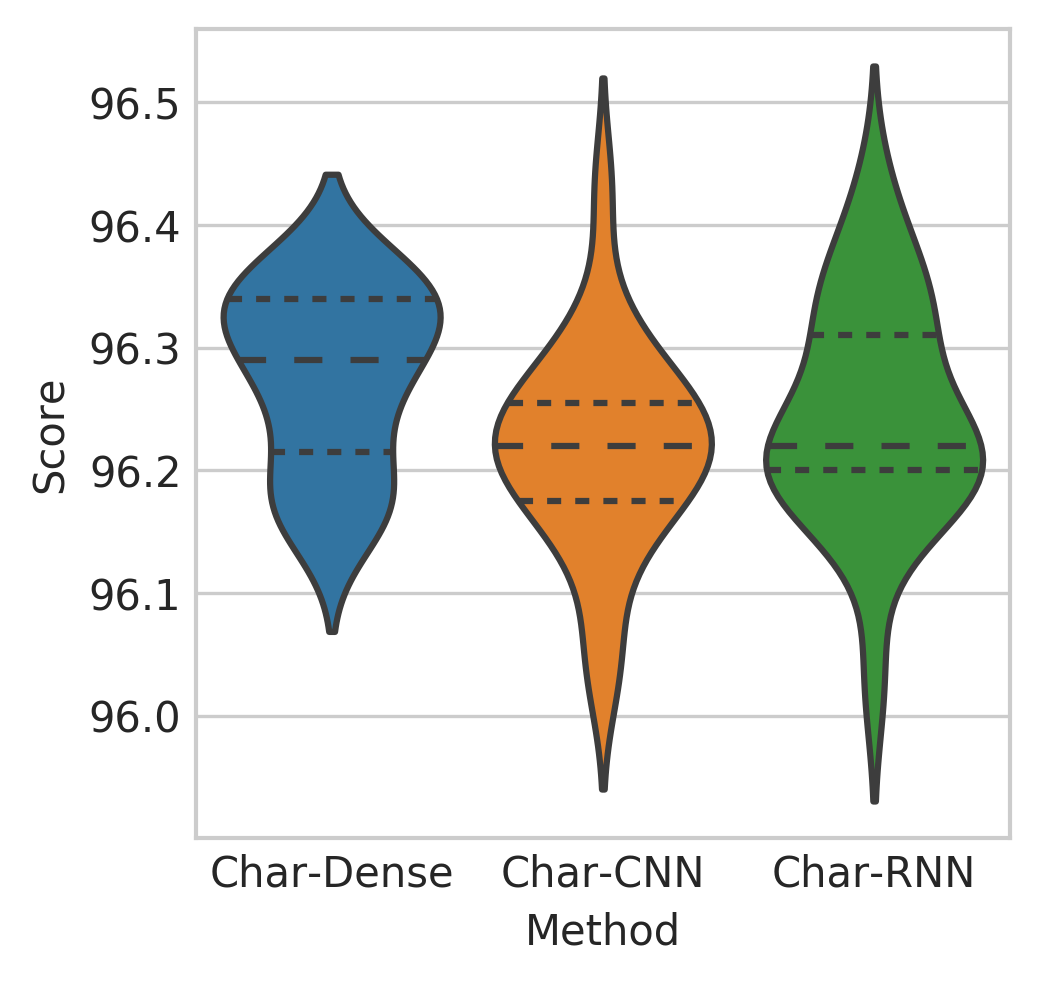}
		\caption{Slot}
		\label{fig:violin_slot}
	\end{subfigure}
	~ 
	\begin{subfigure}[b]{0.3\textwidth}
		\includegraphics[width=\textwidth]{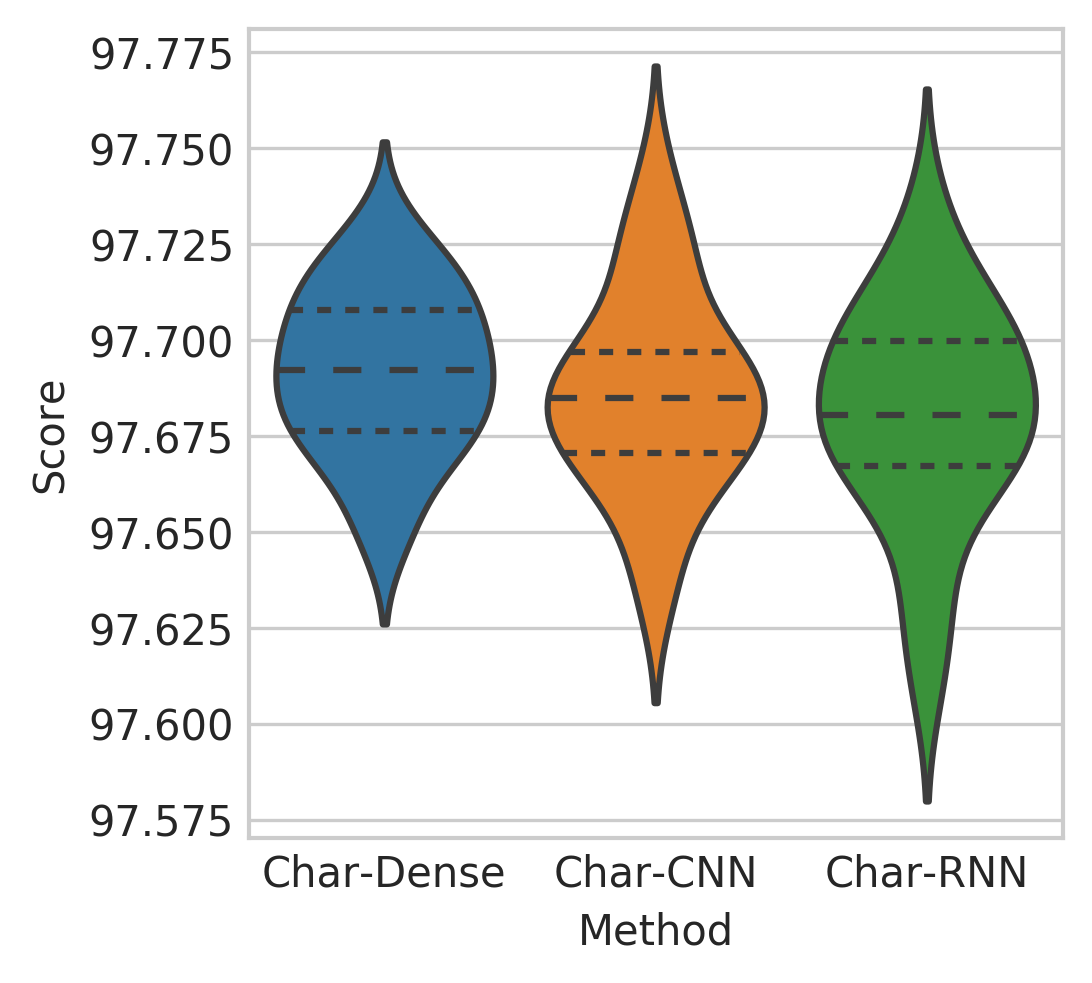}
		\caption{POS}
		\label{fig:violin_pos}
	\end{subfigure}
	~ 
	\begin{subfigure}[b]{0.3\textwidth}
		\includegraphics[width=\textwidth]{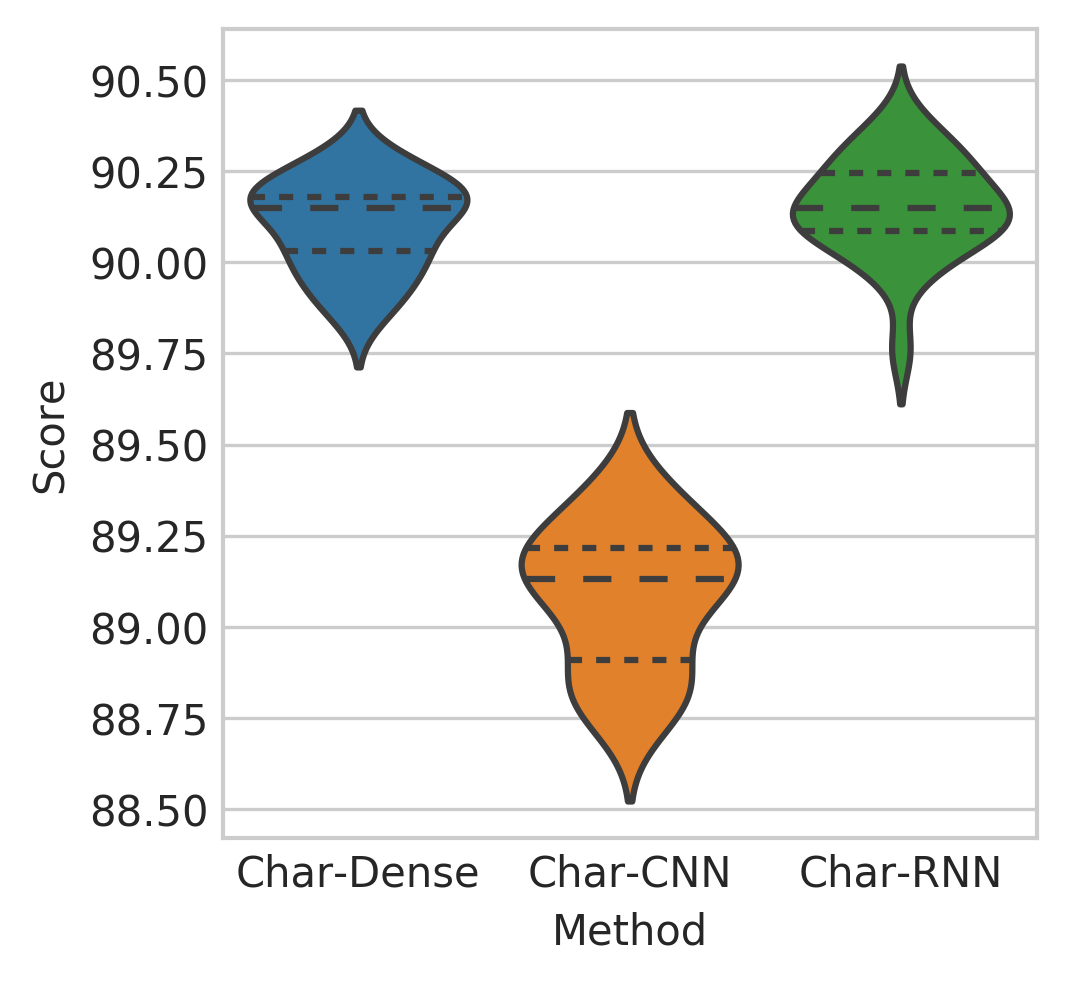}
		\caption{NER}
		\label{fig:violin_ner}
	\end{subfigure}
	\caption{Score distributions for all experiments. Quartiles marked with dashed lines.}\label{fig:violin_plot}
\end{figure}

For a more in-depth analysis of the performance of the proposed method and two baselines, we train each model 20 times with different initial parameters, which are randomly initialized \cite{reimers2017reporting}. Table~\ref{tab:baseline_results} summarizes the mean performance with standard deviation in parentheses. Performance distribution is also visualized using a violin plot in Figure~\ref{fig:violin_plot}.

\subsubsection{Slot Tagging}
On the task of tagging semantic slots using the ATIS dataset, the proposed method shows the best results in terms of both performance and variability. Our method has the highest mean F1-score of 96.28. Furthermore, it has the lowest standard deviation across all runs, which means it is robust to parameter initialization. On the contrary, both CNN and RNN models have lower performance and higher variability compared to the proposed method.\\ 
Analyzing the violin plot reveals that there are also differences in score distribution. While CNN models tend to have a low F1-score on average with occasional high peaks, RNN models have higher F1-score in general but suffer from a large performance drop with poor parameter initialization. This could be one of the reasons why models using CNN seem to have superior performance when only the best performance is reported. On the other hand, our model does not result in peaks or serious drops in performance with different seed values, which makes it more suitable for real-world applications. 

\begin{table}[t]
	\centering
	\begin{tabular}{@{}cccc@{}}
		\toprule
		\multirow{2}{*}{\textbf{Method}} & \multicolumn{3}{c}{\textbf{Task}}                   \\ \cmidrule(l){2-4} 
		& Slot            & POS             & NER             \\ \midrule
		Char-CNN                         & 96.22 (SD 0.08) & 97.68 (SD 0.03) & 89.08 (SD 0.20) \\
		Char-RNN                         & 96.25 (SD 0.09) & 97.68 (SD 0.03) & \textbf{90.15} (SD 0.14) \\
		Char-Dense (Ours)                & \textbf{96.28} (SD \textbf{0.07}) & \textbf{97.69} (SD \textbf{0.02}) & 90.10 (SD \textbf{0.13}) \\ \bottomrule
	\end{tabular}
	\caption{Comparison with baseline models.}
	\label{tab:baseline_results}
\end{table}

\begin{figure}[t]
	\includegraphics[width=0.8\columnwidth]{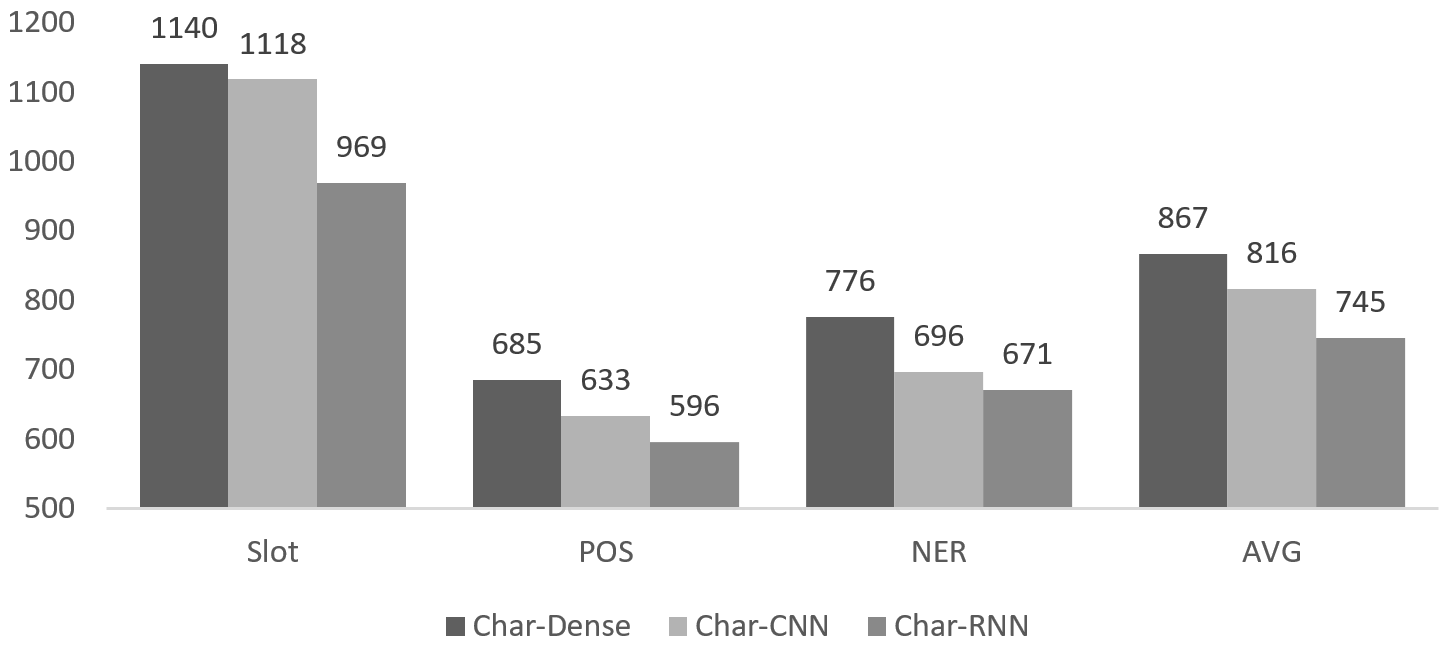}
	\centering
	\caption{Sentence processing speed in terms of number of sentences per second.}
	\label{fig:training_speed}
\end{figure}

\subsubsection{Part-of-Speech Tagging}
The proposed method also achieves the best results on the POS tagging task. Similar to the slot tagging task, our method shows the highest mean accuracy of 97.69 with the lowest standard deviation of 0.02. For the baseline models, CNN and RNN performed on par. \\
CNN-based models have higher variability with high peak performance on this task also, as shown in the violin plot. Similar to the slot tagging task, our method shows the lowest variability, which supports the robustness of this method.

\subsubsection{Named Entity Recognition}
On the NER task, the RNN-based model has a slightly better F1-score (90.15) than the proposed method (90.10). However, our method consistently shows the lowest standard deviation, like as the other tasks, at 0.13. By analyzing the violin plot, we can see that the RNN again shows occasional performance drops for certain cases of poor weight initialization. Unlike the other two tasks, the model utilizing CNN has a relatively poor F1-score and does not show any peaks in performance. 

\subsection{Training Speed}
To compare the efficiency of three models, average training speed (i.e. number of sentences processed per second) is presented in Figure~\ref{fig:training_speed}. All trainings are performed utilizing a single GeForce GTX 1080 Ti GPU, and the RNN model is implemented using the highly efficient cuDNN LSTM API. It is clear that the proposed method has the highest training speed, followed by CNN and RNN. On average, our method was able to process around 867 sentences per second, which is 6.29\% and 16.32\% higher than CNN and RNN, respectively.

\begin{table}[t]
	\centering
	\begin{tabular}{@{}cccccc@{}}
		\toprule
		\multicolumn{2}{c}{\textbf{Slot}}          & \multicolumn{2}{c}{\textbf{POS}}           & \multicolumn{2}{c}{\textbf{NER}}           \\ \midrule
		Approach                  & F1       & Approach                  & Acc.       & Approach                  & F1       \\ \midrule
		\newcite{mesnil2015using}                       & 94.73          & \newcite{toutanova2003feature}                       & 97.24          & 
		\newcite{ando2005framework}                       & 89.31 \\
		\newcite{yao2014spoken}                       & 94.85          & \newcite{manning2011part}               & 97.32          & \newcite{collobert2011natural}                       & 89.59 \\
		\newcite{liu2015recurrent}                       & 94.89          & \newcite{shen2007guided}                & 97.33          & \newcite{huang2015bidirectional}                       & 90.10 \\
		\newcite{yao2014spoken}                       & 95.08          & \newcite{sun2014structure}              & 97.36          & \newcite{chiu2015named}                       & 90.77 \\
		\newcite{peng2015recurrent}                       & 95.25          & \newcite{moore2015improved}             & 97.36          & \newcite{ratinov2009design}                       & 90.80 \\
		\newcite{vu2016bi}                       & 95.56          & \newcite{hajivc2009semi}                & 97.44          & \newcite{lin2009phrase}                       & 90.90 \\
		\newcite{vu2016sequential}                       & 95.61          & \newcite{sogaard2011semisupervised}     & 97.50          & \newcite{passos2014lexicon}                       & 90.90 \\
		\newcite{kurata2016leveraging}                       & 95.66          & 
		\newcite{tsuboi2014neural}              & 97.51          & \newcite{lample2016neural}                       & 90.94 \\
		\newcite{zhu2017encoder}                       & 95.79          & \newcite{huang2015bidirectional}        & 97.55          & \newcite{luo2015joint}                       & 91.20 \\
		\newcite{zhai2017neural}                       & 95.86          & \newcite{choi2016dynamic}                       & 97.64          & \newcite{ma2016end}                       & \textbf{91.21} \\ \midrule
		\makecell{Char-Dense \\w/o CRF (Ours)} & 96.36          & \makecell{Char-Dense \\w/o CRF (Ours)} & \textbf{97.73} & \makecell{Char-Dense \\w/o CRF (Ours)} & 90.28          \\
		\makecell{Char-Dense \\w/ CRF (Ours)}  & \textbf{96.62} & \makecell{Char-Dense \\w/ CRF (Ours)}  & 97.65          & \makecell{Char-Dense \\w/ CRF (Ours)}  & 91.13          \\ \bottomrule
	\end{tabular}
	\caption{Comparison with state-of-the-art approaches in the literature.}
	\label{comparison}
\end{table}

\subsection{Comparison with Published Results}
For comparison with published results, we summarize the performance of our best models along with state-of-the-art approaches in Table~\ref{comparison}. The proposed method was able to surpass the previous state-of-the-art result on the ATIS dataset with a large margin, even without the CRF layer. With the help of CRF, our method obtains a new state-of-the-art result with a 96.62 F1-score. \\
For the POS tagging task with PTB WSJ dataset, we obtain a new state-of-the-art result with a 97.73\% accuracy with the model without a CRF layer. Interestingly, utilizing a CRF layer on this model degraded the performance on this task whereas it helped with the other two tasks. We hypothesize that this is due to the fact that unlike the other two tasks where there are many hard constraints between labels (e.g. an O tag cannot be followed by I- tags), the label dependencies are more "soft" on POS tagging task. In the latter case, it is possible that naively taking label transition probability into account could have a negative impact on performance.\\
On the task of recognizing named entities, we obtain a result that is comparable to state-of-the-art with a 91.13 F1-score when a CRF layer is used. Like in slot tagging task, utilizing CRF lead to a significant increase in performance. It is notable that all results from our method are achieved without depending on any hand-crafted or language/task-specific features (e.g. capitalization, character type, gazetteer), whereas most previous approaches utilizes one or more type of such features. This fact supports the generalizability of the proposed method.

\section{Conclusion and Future Work}
In this paper, we proposed a fast and effective method of using a densely connected network to automatically generate character-level features.  With extensive evaluation, it is shown that this method is robust to parameter initialization and has high processing speed compared to conventional methods such as CNN or RNN. This method has also high generalizability and this is supported by the fact that we were able to obtain superior performance without any task or language specific features.\\
We plan to explore the followings as future work: 1) In this work, we focused on clean text where there are minimal semantic or syntactic errors. We would like to test the robustness of this method against such errors to evaluate whether this method is suitable for real-world applications. 2) Adopting the proposed method and analyzing the effectiveness on other NLP tasks such as neural machine translation or automatic text summarization could also be worth investigating.

\section*{Acknowledgements}

This research was supported by the MSIT (Ministry of Science and ICT), South Korea, under the ITRC (Information Technology Research Center) support program ("Research and Development of Human-Inspired Multiple Intelligence") supervised by the IITP (Institute for Information \& Communications Technology Promotion). Additionally, this work was supported by the National Research Foundation of Korea (NRF) grant funded by the South Korean government (MSIP) (No. NRF-2016R1A2B2015912).

\bibliographystyle{acl}
\bibliography{chanhee2018character}

\begin{thebibliography}{}

\bibitem[\protect\citename{Abadi \bgroup et al.\egroup
  }2016]{abadi2016tensorflow}
Mart{\i}n Abadi, Ashish Agarwal, Paul Barham, Eugene Brevdo, Zhifeng Chen,
  Craig Citro, Greg~S Corrado, Andy Davis, Jeffrey Dean, Matthieu Devin, et~al.
\newblock 2016.
\newblock Tensorflow: Large-scale machine learning on heterogeneous distributed
  systems.
\newblock {\em arXiv preprint arXiv:1603.04467}.

\bibitem[\protect\citename{Ando and Zhang}2005]{ando2005framework}
Rie~Kubota Ando and Tong Zhang.
\newblock 2005.
\newblock A framework for learning predictive structures from multiple tasks
  and unlabeled data.
\newblock {\em Journal of Machine Learning Research}, 6(Nov):1817--1853.

\bibitem[\protect\citename{Chiu and Nichols}2015]{chiu2015named}
Jason~PC Chiu and Eric Nichols.
\newblock 2015.
\newblock Named entity recognition with bidirectional lstm-cnns.
\newblock {\em arXiv preprint arXiv:1511.08308}.

\bibitem[\protect\citename{Choi}2016]{choi2016dynamic}
Jinho~D Choi.
\newblock 2016.
\newblock Dynamic feature induction: The last gist to the state-of-the-art.
\newblock In {\em Proceedings of NAACL-HLT}, pages 271--281.

\bibitem[\protect\citename{Collobert \bgroup et al.\egroup
  }2011]{collobert2011natural}
Ronan Collobert, Jason Weston, L{\'e}on Bottou, Michael Karlen, Koray
  Kavukcuoglu, and Pavel Kuksa.
\newblock 2011.
\newblock Natural language processing (almost) from scratch.
\newblock {\em Journal of Machine Learning Research}, 12(Aug):2493--2537.

\bibitem[\protect\citename{dos Santos and Zadrozny}2014]{dos2014learning}
C{\'\i}cero~Nogueira dos Santos and Bianca Zadrozny.
\newblock 2014.
\newblock Learning character-level representations for part-of-speech tagging.
\newblock In {\em ICML}, pages 1818--1826.

\bibitem[\protect\citename{Finkel \bgroup et al.\egroup
  }2005]{finkel2005exploring}
Jenny Finkel, Shipra Dingare, Christopher~D Manning, Malvina Nissim, Beatrice
  Alex, and Claire Grover.
\newblock 2005.
\newblock Exploring the boundaries: gene and protein identification in
  biomedical text.
\newblock {\em BMC bioinformatics}, 6(1):S5.

\bibitem[\protect\citename{Gal and Ghahramani}2016]{gal2016theoretically}
Yarin Gal and Zoubin Ghahramani.
\newblock 2016.
\newblock A theoretically grounded application of dropout in recurrent neural
  networks.
\newblock In {\em Advances in neural information processing systems}, pages
  1019--1027.

\bibitem[\protect\citename{Graves and Schmidhuber}2005]{graves2005framewise}
Alex Graves and J{\"u}rgen Schmidhuber.
\newblock 2005.
\newblock Framewise phoneme classification with bidirectional lstm and other
  neural network architectures.
\newblock {\em Neural Networks}, 18(5):602--610.

\bibitem[\protect\citename{Haji{\v{c}} \bgroup et al.\egroup
  }2009]{hajivc2009semi}
Jan Haji{\v{c}}, Jan Raab, Miroslav Spousta, et~al.
\newblock 2009.
\newblock Semi-supervised training for the averaged perceptron pos tagger.
\newblock In {\em Proceedings of the 12th Conference of the European Chapter of
  the Association for Computational Linguistics}, pages 763--771. Association
  for Computational Linguistics.

\bibitem[\protect\citename{He \bgroup et al.\egroup }2016]{he2016deep}
Kaiming He, Xiangyu Zhang, Shaoqing Ren, and Jian Sun.
\newblock 2016.
\newblock Deep residual learning for image recognition.
\newblock In {\em Proceedings of the IEEE conference on computer vision and
  pattern recognition}, pages 770--778.

\bibitem[\protect\citename{Hochreiter and Schmidhuber}1997]{hochreiter1997long}
Sepp Hochreiter and J{\"u}rgen Schmidhuber.
\newblock 1997.
\newblock Long short-term memory.
\newblock {\em Neural computation}, 9(8):1735--1780.

\bibitem[\protect\citename{Huang \bgroup et al.\egroup
  }2015]{huang2015bidirectional}
Zhiheng Huang, Wei Xu, and Kai Yu.
\newblock 2015.
\newblock Bidirectional lstm-crf models for sequence tagging.
\newblock {\em arXiv preprint arXiv:1508.01991}.

\bibitem[\protect\citename{Keskar \bgroup et al.\egroup }2016]{keskar2016large}
Nitish~Shirish Keskar, Dheevatsa Mudigere, Jorge Nocedal, Mikhail Smelyanskiy,
  and Ping Tak~Peter Tang.
\newblock 2016.
\newblock On large-batch training for deep learning: Generalization gap and
  sharp minima.
\newblock {\em arXiv preprint arXiv:1609.04836}.

\bibitem[\protect\citename{Kim and Snyder}2012]{kim2012universal}
Young-Bum Kim and Benjamin Snyder.
\newblock 2012.
\newblock Universal grapheme-to-phoneme prediction over latin alphabets.
\newblock In {\em Proceedings of the 2012 Joint Conference on Empirical Methods
  in Natural Language Processing and Computational Natural Language Learning},
  pages 332--343. Association for Computational Linguistics.

\bibitem[\protect\citename{Kim \bgroup et al.\egroup }2011]{kim2011universal}
Young-Bum Kim, Jo{\~a}o~V Gra{\c{c}}a, and Benjamin Snyder.
\newblock 2011.
\newblock Universal morphological analysis using structured nearest neighbor
  prediction.
\newblock In {\em Proceedings of the Conference on Empirical Methods in Natural
  Language Processing}, pages 322--332. Association for Computational
  Linguistics.

\bibitem[\protect\citename{Kim \bgroup et al.\egroup }2014]{kim2014training}
Young-Bum Kim, Heemoon Chae, Benjamin Snyder, and Yu-Seop Kim.
\newblock 2014.
\newblock Training a korean srl system with rich morphological features.
\newblock In {\em Proceedings of the 52nd Annual Meeting of the Association for
  Computational Linguistics (Volume 2: Short Papers)}, volume~2, pages
  637--642.

\bibitem[\protect\citename{Kim \bgroup et al.\egroup }2015]{kim2015new}
Young-Bum Kim, Karl Stratos, Ruhi Sarikaya, and Minwoo Jeong.
\newblock 2015.
\newblock New transfer learning techniques for disparate label sets.
\newblock In {\em Proceedings of the 53rd Annual Meeting of the Association for
  Computational Linguistics and the 7th International Joint Conference on
  Natural Language Processing (Volume 1: Long Papers)}, volume~1, pages
  473--482.

\bibitem[\protect\citename{Kim \bgroup et al.\egroup
  }2016]{kim2016frustratingly}
Young-Bum Kim, Karl Stratos, and Ruhi Sarikaya.
\newblock 2016.
\newblock Frustratingly easy neural domain adaptation.
\newblock In {\em Proceedings of COLING 2016, the 26th International Conference
  on Computational Linguistics: Technical Papers}, pages 387--396.

\bibitem[\protect\citename{Kingma and Ba}2014]{kingma2014adam}
Diederik Kingma and Jimmy Ba.
\newblock 2014.
\newblock Adam: A method for stochastic optimization.
\newblock {\em arXiv preprint arXiv:1412.6980}.

\bibitem[\protect\citename{Kurata \bgroup et al.\egroup
  }2016]{kurata2016leveraging}
Gakuto Kurata, Bing Xiang, Bowen Zhou, and Mo~Yu.
\newblock 2016.
\newblock Leveraging sentence-level information with encoder lstm for semantic
  slot filling.
\newblock {\em arXiv preprint arXiv:1601.01530}.

\bibitem[\protect\citename{Lample \bgroup et al.\egroup
  }2016]{lample2016neural}
Guillaume Lample, Miguel Ballesteros, Sandeep Subramanian, Kazuya Kawakami, and
  Chris Dyer.
\newblock 2016.
\newblock Neural architectures for named entity recognition.
\newblock {\em arXiv preprint arXiv:1603.01360}.

\bibitem[\protect\citename{Lin and Wu}2009]{lin2009phrase}
Dekang Lin and Xiaoyun Wu.
\newblock 2009.
\newblock Phrase clustering for discriminative learning.
\newblock In {\em Proceedings of the Joint Conference of the 47th Annual
  Meeting of the ACL and the 4th International Joint Conference on Natural
  Language Processing of the AFNLP: Volume 2-Volume 2}, pages 1030--1038.
  Association for Computational Linguistics.

\bibitem[\protect\citename{Liu and Lane}2015]{liu2015recurrent}
Bing Liu and Ian Lane.
\newblock 2015.
\newblock Recurrent neural network structured output prediction for spoken
  language understanding.
\newblock In {\em Proc. NIPS Workshop on Machine Learning for Spoken Language
  Understanding and Interactions}.

\bibitem[\protect\citename{Luo \bgroup et al.\egroup }2015]{luo2015joint}
Gang Luo, Xiaojiang Huang, Chin-Yew Lin, and Zaiqing Nie.
\newblock 2015.
\newblock Joint entity recognition and disambiguation.
\newblock In {\em Proceedings of the 2015 Conference on Empirical Methods in
  Natural Language Processing}, pages 879--888.

\bibitem[\protect\citename{Ma and Hovy}2016]{ma2016end}
Xuezhe Ma and Eduard Hovy.
\newblock 2016.
\newblock End-to-end sequence labeling via bi-directional lstm-cnns-crf.
\newblock {\em arXiv preprint arXiv:1603.01354}.

\bibitem[\protect\citename{Manning}2011]{manning2011part}
Christopher~D Manning.
\newblock 2011.
\newblock Part-of-speech tagging from 97\% to 100\%: is it time for some
  linguistics?
\newblock In {\em International Conference on Intelligent Text Processing and
  Computational Linguistics}, pages 171--189. Springer.

\bibitem[\protect\citename{Marcus \bgroup et al.\egroup
  }1993]{marcus1993building}
Mitchell~P Marcus, Mary~Ann Marcinkiewicz, and Beatrice Santorini.
\newblock 1993.
\newblock Building a large annotated corpus of english: The penn treebank.
\newblock {\em Computational linguistics}, 19(2):313--330.

\bibitem[\protect\citename{Mesnil \bgroup et al.\egroup }2015]{mesnil2015using}
Gr{\'e}goire Mesnil, Yann Dauphin, Kaisheng Yao, Yoshua Bengio, Li~Deng, Dilek
  Hakkani-Tur, Xiaodong He, Larry Heck, Gokhan Tur, Dong Yu, et~al.
\newblock 2015.
\newblock Using recurrent neural networks for slot filling in spoken language
  understanding.
\newblock {\em IEEE/ACM Transactions on Audio, Speech, and Language
  Processing}, 23(3):530--539.

\bibitem[\protect\citename{Moore}2015]{moore2015improved}
Robert Moore.
\newblock 2015.
\newblock An improved tag dictionary for faster part-of-speech tagging.
\newblock In {\em Proc. of EMNLP}. Citeseer.

\bibitem[\protect\citename{Passos \bgroup et al.\egroup
  }2014]{passos2014lexicon}
Alexandre Passos, Vineet Kumar, and Andrew McCallum.
\newblock 2014.
\newblock Lexicon infused phrase embeddings for named entity resolution.
\newblock {\em arXiv preprint arXiv:1404.5367}.

\bibitem[\protect\citename{Peng and Yao}2015]{peng2015recurrent}
Baolin Peng and Kaisheng Yao.
\newblock 2015.
\newblock Recurrent neural networks with external memory for language
  understanding.
\newblock {\em arXiv preprint arXiv:1506.00195}.

\bibitem[\protect\citename{Pennington \bgroup et al.\egroup
  }2014]{pennington2014glove}
Jeffrey Pennington, Richard Socher, and Christopher Manning.
\newblock 2014.
\newblock Glove: Global vectors for word representation.
\newblock In {\em Proceedings of the 2014 conference on empirical methods in
  natural language processing (EMNLP)}, pages 1532--1543.

\bibitem[\protect\citename{Ratinov and Roth}2009]{ratinov2009design}
Lev Ratinov and Dan Roth.
\newblock 2009.
\newblock Design challenges and misconceptions in named entity recognition.
\newblock In {\em Proceedings of the Thirteenth Conference on Computational
  Natural Language Learning}, pages 147--155. Association for Computational
  Linguistics.

\bibitem[\protect\citename{Reimers and Gurevych}2017]{reimers2017reporting}
Nils Reimers and Iryna Gurevych.
\newblock 2017.
\newblock Reporting score distributions makes a difference: Performance study
  of lstm-networks for sequence tagging.
\newblock {\em arXiv preprint arXiv:1707.09861}.

\bibitem[\protect\citename{Sarikaya and Deng}2007]{sarikaya2007joint}
Ruhi Sarikaya and Yonggang Deng.
\newblock 2007.
\newblock Joint morphological-lexical language modeling for machine
  translation.
\newblock In {\em Human Language Technologies 2007: The Conference of the North
  American Chapter of the Association for Computational Linguistics; Companion
  Volume, Short Papers}, pages 145--148. Association for Computational
  Linguistics.

\bibitem[\protect\citename{Sennrich \bgroup et al.\egroup
  }2015]{sennrich2015neural}
Rico Sennrich, Barry Haddow, and Alexandra Birch.
\newblock 2015.
\newblock Neural machine translation of rare words with subword units.
\newblock {\em arXiv preprint arXiv:1508.07909}.

\bibitem[\protect\citename{Shen \bgroup et al.\egroup }2007]{shen2007guided}
Libin Shen, Giorgio Satta, and Aravind Joshi.
\newblock 2007.
\newblock Guided learning for bidirectional sequence classification.
\newblock In {\em ACL}, volume~7, pages 760--767. Citeseer.

\bibitem[\protect\citename{Smith \bgroup et al.\egroup }2017]{smith2017don}
Samuel~L Smith, Pieter-Jan Kindermans, and Quoc~V Le.
\newblock 2017.
\newblock Don't decay the learning rate, increase the batch size.
\newblock {\em arXiv preprint arXiv:1711.00489}.

\bibitem[\protect\citename{S{\o}gaard}2011]{sogaard2011semisupervised}
Anders S{\o}gaard.
\newblock 2011.
\newblock Semisupervised condensed nearest neighbor for part-of-speech tagging.
\newblock In {\em Proceedings of the 49th Annual Meeting of the Association for
  Computational Linguistics: Human Language Technologies: short papers-Volume
  2}, pages 48--52. Association for Computational Linguistics.

\bibitem[\protect\citename{Srivastava \bgroup et al.\egroup
  }2014]{srivastava2014dropout}
Nitish Srivastava, Geoffrey Hinton, Alex Krizhevsky, Ilya Sutskever, and Ruslan
  Salakhutdinov.
\newblock 2014.
\newblock Dropout: A simple way to prevent neural networks from overfitting.
\newblock {\em The Journal of Machine Learning Research}, 15(1):1929--1958.

\bibitem[\protect\citename{Sun}2014]{sun2014structure}
Xu~Sun.
\newblock 2014.
\newblock Structure regularization for structured prediction.
\newblock In {\em Advances in Neural Information Processing Systems}, pages
  2402--2410.

\bibitem[\protect\citename{Tjong Kim~Sang and
  De~Meulder}2003]{tjong2003introduction}
Erik~F Tjong Kim~Sang and Fien De~Meulder.
\newblock 2003.
\newblock Introduction to the conll-2003 shared task: Language-independent
  named entity recognition.
\newblock In {\em Proceedings of the seventh conference on Natural language
  learning at HLT-NAACL 2003-Volume 4}, pages 142--147. Association for
  Computational Linguistics.

\bibitem[\protect\citename{Toutanova \bgroup et al.\egroup
  }2003]{toutanova2003feature}
Kristina Toutanova, Dan Klein, Christopher~D Manning, and Yoram Singer.
\newblock 2003.
\newblock Feature-rich part-of-speech tagging with a cyclic dependency network.
\newblock In {\em Proceedings of the 2003 Conference of the North American
  Chapter of the Association for Computational Linguistics on Human Language
  Technology-Volume 1}, pages 173--180. Association for Computational
  Linguistics.

\bibitem[\protect\citename{Tsuboi}2014]{tsuboi2014neural}
Yuta Tsuboi.
\newblock 2014.
\newblock Neural networks leverage corpus-wide information for part-of-speech
  tagging.
\newblock In {\em EMNLP}, pages 938--950.

\bibitem[\protect\citename{Vu \bgroup et al.\egroup }2016]{vu2016bi}
Ngoc~Thang Vu, Pankaj Gupta, Heike Adel, and Hinrich Sch{\"u}tze.
\newblock 2016.
\newblock Bi-directional recurrent neural network with ranking loss for spoken
  language understanding.
\newblock In {\em Acoustics, Speech and Signal Processing (ICASSP), 2016 IEEE
  International Conference on}, pages 6060--6064. IEEE.

\bibitem[\protect\citename{Vu}2016]{vu2016sequential}
Ngoc~Thang Vu.
\newblock 2016.
\newblock Sequential convolutional neural networks for slot filling in spoken
  language understanding.
\newblock {\em arXiv preprint arXiv:1606.07783}.

\bibitem[\protect\citename{Yao \bgroup et al.\egroup }2014]{yao2014spoken}
Kaisheng Yao, Baolin Peng, Yu~Zhang, Dong Yu, Geoffrey Zweig, and Yangyang Shi.
\newblock 2014.
\newblock Spoken language understanding using long short-term memory neural
  networks.
\newblock In {\em Spoken Language Technology Workshop (SLT), 2014 IEEE}, pages
  189--194. IEEE.

\bibitem[\protect\citename{Zhai \bgroup et al.\egroup }2017]{zhai2017neural}
Feifei Zhai, Saloni Potdar, Bing Xiang, and Bowen Zhou.
\newblock 2017.
\newblock Neural models for sequence chunking.

\bibitem[\protect\citename{Zhu and Yu}2017]{zhu2017encoder}
Su~Zhu and Kai Yu.
\newblock 2017.
\newblock Encoder-decoder with focus-mechanism for sequence labelling based
  spoken language understanding.
\newblock In {\em Acoustics, Speech and Signal Processing (ICASSP), 2017 IEEE
  International Conference on}, pages 5675--5679. IEEE.

\end{thebibliography}

\clearpage
\begin{appendices}
\section{Hyperparameters}
\label{appendix:hyperparameters}

\begin{table}[h]
	\centering
	\begin{tabular}{@{}ccccc@{}}
		\toprule
		\textbf{Group}                            & \textbf{Hyperparameter}     & \textbf{Slot} & \textbf{POS} & \textbf{NER} \\ \midrule
		\multirow{3}{*}{Char-CNN}                 & Window size                 & 3             & 3            & 3            \\
		& Number of filters           & 30            & 30           & 30           \\
		& Character dimension         & 30            & 30           & 30           \\ \midrule
		\multirow{2}{*}{Char-RNN}                 & Layer size                  & 50            & 50           & 50           \\
		& Character dimension         & 50            & 50           & 50           \\ \midrule
		\multirow{2}{*}{Char-Dense}               & Layer size                  & 50            & 50           & 50           \\
		& Number of pieces per word   & 2             & 2            & 2            \\ \midrule
		\multirow{5}{*}{Word-level}               & Pre-trained word embeddings & GloVe 300d    & GloVe 300d   & GloVe 300d   \\
		& RNN layer size              & 350           & 350          & 350          \\
		& RNN layer depth             & 2             & 3            & 3            \\
		& Pre-RNN layer size          & 350           & 350          & None         \\
		& Post-RNN layer size         & 350           & 350          & None         \\ \midrule
		\multirow{4}{*}{Dropout keep probability} & Char-Dense                  & 0.7           & 0.7          & 0.7          \\
		& Word feature                & 0.9           & 0.9          & 0.9          \\
		& Word-level RNN layer        & 0.5           & 0.5          & 0.5          \\
		& Pre/post-RNN layers         & 0.5           & 0.5          & 0.5          \\ \midrule
		\multirow{2}{*}{Training}                 & Initial batch size          & 8             & 16           & 16           \\
		& Number of epochs            & 100           & 100          & 100          \\ \bottomrule
	\end{tabular}
	\caption{Chosen hyperparameters for all experiments.}
	\label{hyperparameters}
\end{table}

\end{appendices}

\end{document}